\documentclass[letterpaper]{article} % DO NOT CHANGE THIS
\usepackage{aaai24}  % DO NOT CHANGE THIS
\usepackage{times}  % DO NOT CHANGE THIS
\usepackage{helvet}  % DO NOT CHANGE THIS
\usepackage{courier}  % DO NOT CHANGE THIS
\usepackage[hyphens]{url}  % DO NOT CHANGE THIS
\usepackage{graphicx} % DO NOT CHANGE THIS
\usepackage{natbib}  % DO NOT CHANGE THIS AND DO NOT ADD ANY OPTIONS TO IT
\usepackage{caption} % DO NOT CHANGE THIS AND DO NOT ADD ANY OPTIONS TO IT
\usepackage{algorithm}
\usepackage{algorithmic}
\usepackage{algorithm}
\usepackage{algorithmic}
\usepackage{amsfonts}

\usepackage{cite}
\usepackage{multirow}
\usepackage{array}
\usepackage{hhline}
\usepackage{makecell}
\usepackage{booktabs}
\usepackage{amsmath}
\usepackage{amssymb}

\usepackage[capitalize]{cleveref}
\crefname{section}{Sec.}{Secs.}
\Crefname{section}{Section}{Sections}
\Crefname{table}{Table}{Tables}
\crefname{table}{Tab.}{Tabs.}
\Crefname{equation}{Equation}{Equations}
\crefname{equation}{Eqn.}{Eqns.}

\usepackage{xcolor}
\newcommand{\lxm}[1]{\textcolor[rgb]{0,0,0}{#1}}
\newcommand{\lxmup}[1]{\textcolor[rgb]{0,0,0}{#1}}

\newcommand{\zyb}[1]{\textcolor[rgb]{0,0,0}{#1}}
\newcommand{\etal}{\textit{et al}.}
\newcommand{\ie}{\textit{i}.\textit{e}.}
\newcommand{\eg}{\textit{e}.\textit{g}.}

\usepackage{newfloat}
\usepackage{listings}
\DeclareCaptionStyle{ruled}{labelfont=normalfont,labelsep=colon,strut=off} % DO NOT CHANGE THIS
\floatstyle{ruled}
\newfloat{listing}{tb}{lst}{}
\floatname{listing}{Listing}
\title{VQ-Font: Few-Shot Font Generation with Structure-Aware Enhancement and Quantization}
\author {
    Mingshuai Yao\textsuperscript{\rm 1}, 
    Yabo Zhang\textsuperscript{\rm 1},
    Xianhui Lin\textsuperscript{} ,
    Xiaoming Li\textsuperscript{\rm 2},
    Wangmeng Zuo\textsuperscript{\rm 1}\thanks{Corresponding Author}
}
\affiliations {
    \textsuperscript{\rm 1} Harbin Institute of Technology \\ %\quad
    \textsuperscript{\rm 2} Nanyang Technological University\\
    \{ymsoyosmy,hitzhangyabo2017,xhlin129,csxmli\}@gmail.com, wmzuo@hit.edu.cn
}
\usepackage{bibentry}
\usepackage{amsmath,amsfonts,bm}

\def\eqref#1{equation~\ref{#1}}
\def\1{\bm{1}}

\def\vf{{\bm{f}}}

\def\vp{{\bm{p}}}

\def\vs{{\bm{s}}}

\def\mA{{\bm{A}}}

\def\mI{{\bm{I}}}

\def\mW{{\bm{W}}}

\def\mZ{{\bm{Z}}}

\DeclareMathAlphabet{\mathsfit}{\encodingdefault}{\sfdefault}{m}{sl}
\SetMathAlphabet{\mathsfit}{bold}{\encodingdefault}{\sfdefault}{bx}{n}

\begin{document}

\maketitle

% \if 0
 \begin{abstract}

Few-shot font generation is challenging, as it needs to capture the fine-grained stroke styles from a \lxm{limited set of} reference glyphs, and then \lxm{transfer} to other characters, which are expected to \lxm{have similar styles}. However, due to the diversity and complexity of Chinese \lxm{font styles}, the synthesized glyphs of existing methods usually exhibit visible artifacts, such as missing details and distorted strokes. 
In this paper, we propose a VQGAN-based framework (\ie, VQ-Font) to enhance glyph fidelity through token prior \lxm{refinement} and structure-aware \lxm{enhancement}.
Specifically, we pre-train a VQGAN 
%via self-reconstruction 
\lxm{to encapsulate font token} {prior} within a codebook. Subsequently, VQ-Font refines the synthesized {glyphs} \lxm{with the codebook to eliminate the domain gap between synthesized and real-world strokes.}
%by mapping the generated strokes to the high-quality font structure priors.
%
Furthermore, our VQ-Font leverages the inherent design of Chinese characters, where structure components such as radicals and character components are combined in specific arrangements, to recalibrate fine-grained styles based on references. This process improves the matching and fusion of styles at the structure level.
Both modules collaborate to enhance the fidelity of the generated fonts.
Experiments on a collected font dataset show that our VQ-Font outperforms the competing methods both quantitatively and qualitatively, especially in generating \lxm{challenging styles}.
Code is available at https://github.com/Yaomingshuai/VQ-Font.
\end{abstract}

\section{Introduction}
\begin{figure}[t]
\centering
\includegraphics[width=0.99\columnwidth]{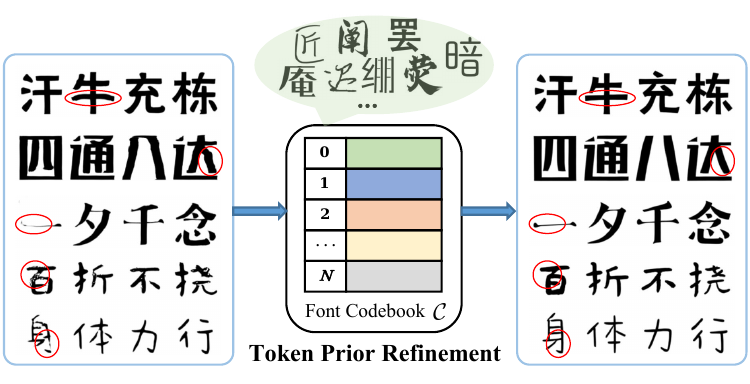} % Reduce the figure size so that it is slightly narrower than the column. Don't use precise values for figure width.This setup will avoid overfull boxes.
    %\put(0,0){\small{Input}}
    %\vspace{1mm}
%\caption{\textbf{The process we propose for recovering font details using a codebook.}}
\caption{\lxm{\textbf{Left: }The illustration of missing details and distorted strokes, and \textbf{Right: }the refinement process through token prior encapsulated in our codebook.}}
\label{introduction}
\end{figure} 

%背景介绍，说明font library很重要，且设计font非常费时费力
Font library elegantly represents text information in computer systems and has tremendous value in commercial and artistic applications.
% Manually designing such a font library is extremely time-consuming, especially for logographic languages containing thousands of characters.
% As representations of text, fonts have rich application scenarios and wide commercial values. 
%
Manually designing such a library is highly resource-intensive and laborious, especially for logographic languages containing thousands of characters (\emph{e.g.}, Chinese, Japanese, and Korean).
%
%Fortunately, considering that the designing process of most glyphs is repetitive, it is feasible to create a new glyph by directly borrowing styles from other designed glyphs at different levels of granularity, \emph{e.g.,} structure and stroke levels.
\lxm{However, each glyph is typically constructed using fundamental strokes, thereby making it feasible to create a new glyph by directly adopting styles from other reference glyphs at different levels of granularity, \emph{e.g.,} structure and stroke. }
%
% In general, when designing a new glyph for a character, calligraphers 
% each glyph in font library is independently designed by professional calligraphers, 

% A font library consists of glyphs corresponding to the characters, where each of them  is carefully designed by professional calligraphers.
%
% Nonetheless, logographic languages typically contains thousands of characters (\emph{e.g.}, Chinese, Japanese, and Korean), so designing a new style of font library is highly resource-intensive and laborious.
%
% Different styles of fonts are usually required in different situations to convey a specific message.
% For example, formal print fonts are needed in textbooks, while attractive artistic fonts are needed in advertising slogans. 
% However, designing a font library is a very time-consuming and labor-intensive process, especially for  Chinese that has over 60k characters. 
% Designers not only need to consider the overall structure of Chinese characters to ensure that the content is not destroyed, but also need to take into account fine-grained details such as stroke characteristics and edge details to ensure consistency in style.\par

%动机1：开头直接提出Few-shot font generation是解决上述“费时费力”的问题的任务，再叙述之前方法，顺便引出FsFont，提出动机1.
%Few-shot font generation is a common and effective technique to reduce the high cost of creating a new style of font library.
\lxm{Among these recent methods, few-shot font generation attracts significant attention, {as it is effective in reducing the human labor required for designing a target font library.}}
%
% Existing methods formulate it as an image-to-image translation problem.
%
%Given a few glyphs as style references, they translate a content glyph to a target one by adaptively querying fine-grained styles from references.
\lxm{Given limited target glyphs as style references, they usually translate a content glyph to a target one by adaptively querying fine-grained styles from references.}
\lxm{FS-Font~\cite{tang2022few}, one of the most representative methods, {proposes a style aggregation module based on cross-attention mechanism.} It aggregates patch-grained styles to obtain target glyphs. However, it's noteworthy that the Chinese comprise over 3,000 characters, and subtle variations in strokes can give rise to completely different characters. 
 % (\emph{e.g.}, `\begin{CJK*}{UTF8}{bsmi}已\end{CJK*}' and `\begin{CJK*}{UTF8}{bsmi}己\end{CJK*}')
Besides, nearly all the existing methods learn the representations of characters from scratch, inevitably leading to issues of missing or distorted strokes (see the left part in~\cref{introduction}).
Their performance encounters further degradation when handling complex styles, \eg, serif and artistic fonts.}
%Owing to the intricate nature of Chinese characters and their vast diversity, all the existing methods, including Fs-Font, learning the representations of characters from scratch, often suffer from missing strokes, distorted strokes, and loss of fine details, particularly in serif and artistic fonts.
% generate font images directly in continuous space, ignoring the discrete and high-frequency information of Chinese characters. As a result, they 
% often suffer from missing strokes, distorted strokes, and loss of fine details, particularly in serif and artistic fonts.

%
% Albeit considerable synthesized glyphs
% , thus learning fine-grained styles.
% zi2zi ~\cite{tian2017master} using image-to-image translation networks similar to Pixel2Pixel ~\cite{isola2017image}, achieves font generation by learning the mapping function between different fonts, but it can only generate fonts that have been seen during training. 
% However, disentanglement-based methods typically perform an average operations on the extracted features. In addition, due to the high entanglement between the content and style information in font images, these methods tend to ignore certain local details of fonts.
%动机2 感觉可以和动机1合在一起
%%%
% 新开一段分析字体模糊，缺少笔画的原因：
% 1. 空间连续，忽略低频信息，方便下面引出vqgan
% 2. attention对应的过程有噪声信息，方便引出结构reweight
%%%

\begin{figure}[t]
\centering
\includegraphics[width=0.99\columnwidth]{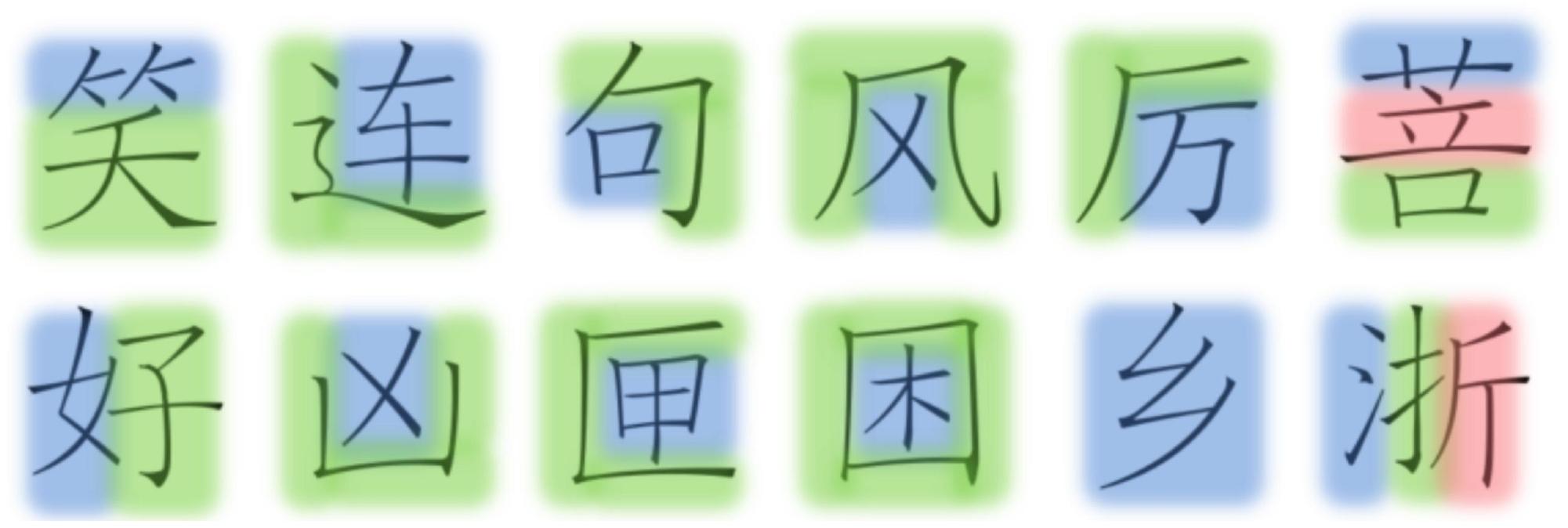} % Reduce the figure size so that it is slightly narrower than the column. Don't use precise values for figure width.This setup will avoid overfull boxes.
    %\put(0,0){\small{Input}}
    %\vspace{1mm}
\caption{\textbf{Structure component divisions in Chinese characters.} Different colors represent different components.}
\label{12_stru}
\end{figure}

%动机2：先提出FsFont只是内容字和风格字patch-level的对应，缺少对整体结构的关注。
Furthermore, FS-Font takes the patch tokens from content glyphs as queries while those from reference glyphs act as keys and values. This approach focuses on the patch-level correspondence between content and reference glyphs, which neglects the inherent principles of character design. As an ideographic writing system, Chinese characters carry distinct structure information. As shown in~\cref{12_stru}, nearly all Chinese characters can be divided into 12 structures, such as top-bottom arrangement and left-right arrangement. Among them, most structures contain two or three components. According to the structure division of Chinese characters, when the radical or character component of the content glyphs appears in the reference glyphs, we aim to treat these structure components as unified entities rather than focusing only on patch-level correspondence.\par

% \begin{figure}[t]
% \centering\includegraphics[width=3cm]{example-grid-100x100pt}
% %\includegraphics[width=0.9\columnwidth]{introduction.png} % Reduce the figure size so that it is slightly narrower than the column. Don't use precise values for figure width.This setup will avoid overfull boxes.
%     %\put(0,0){\small{Input}}
%      %\vspace{1mm}
% \caption{\textbf{The process we propose for recovering font details using a codebook.}}
% \label{introduction}
% \end{figure} 

\lxm{To improve the fidelity of synthesized glyphs, we propose VQ-Font, a framework encompassing the structure-aware enhancement and token prior refinement.
To be specific, we firstly pre-train a VQGAN model~\cite{esser2021taming} on diverse and high-quality font images. This VQGAN model has the ability to generate font images aligning well with the real-world manifold. Then, we employ a Transformer~\cite{vaswani2017attention} to globally model the synthesized font images and predict their corresponding indices within our pre-trained codebook, which can refine the font images by mapping into the token prior space.
In addition, we propose a {Structure-level Style Enhancement Module~(SSEM)} to explicitly incorporate Chinese character structure information. By establishing a correspondence between the structure components of the content and reference glyphs, it recalibrates the fine-grained styles derived from the references, thereby facilitating the accurate learning and matching process of glyph style transformation.}

In summary, our work has three main contributions:
\begin{itemize}
    %\item we introduce a learned codebook with font detail priors for few-shot font generation for the first time. By matching the discrete space in font codebook, we greatly alleviate the issue of missing details and distorted strokes in previous methods.
    \item \lxm{We introduce a font codebook that encapsulates token prior to refine synthesized font images. By mapping the synthesized font into the discrete space defined by the codebook, our VQ-Font can effectively address the issues of missing details and distorted strokes.}
    \item We explicitly incorporate the \lxm{design criterion} of Chinese characters by introducing structure-level correspondence. This promotes the model to better learn the styles of the reference glyphs at the structure level.
    %\item Our VQ-Font is capable of generating more complex fonts and outperforms previous methods in terms of both quantity and quality.
    \item \lxm{Our VQ-Font outperforms these competing methods in both quantitative and qualitative evaluation. It is also capable of generating complex styles with better fidelity.}
\end{itemize}

\begin{figure*}[t]
\vspace{-8pt}
\setlength{\belowcaptionskip}{-5pt}
\centering
\includegraphics[width=0.98\textwidth]{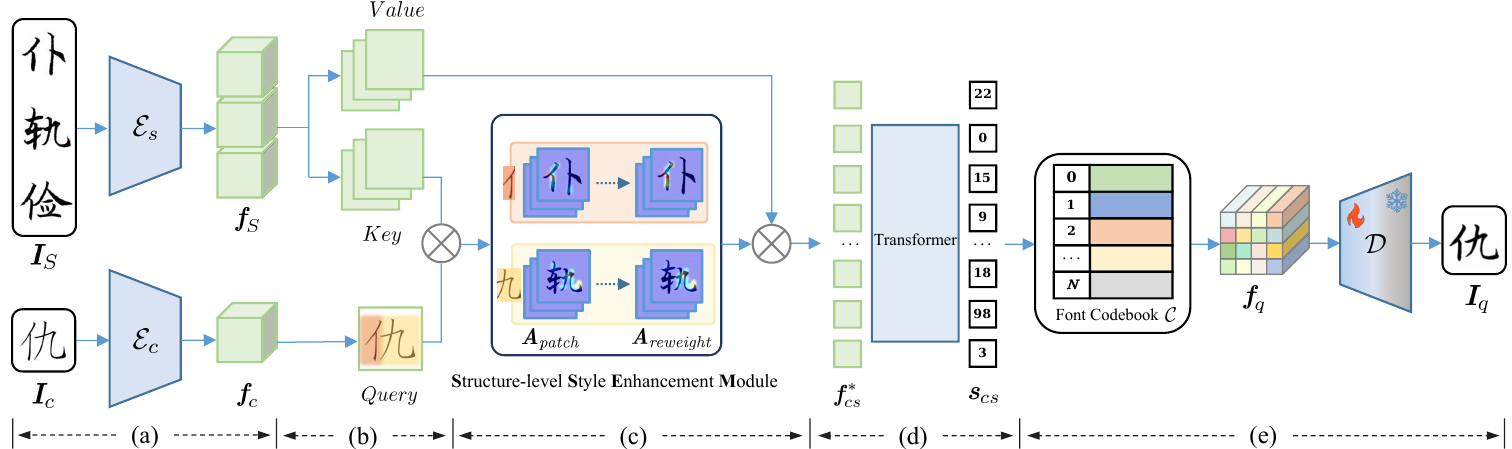} % Reduce the figure size so that it is slightly narrower than the column.
\caption{\lxmup{\textbf{Overview of VQ-Font framework. }VQ-Font mainly consists of the following components: a style encoder $\mathcal{E}_s$, a content encoder $\mathcal{E}_c$, a cross-attention module, a structure-level style enhancement module, a Transformer module, a pre-trained font codebook $\mathcal{C}$ and font decoder $\mathcal{D}$. \textbf{(a)} $\mathcal{E}_s$ and $\mathcal{E}_c$ first map $\mI_S$ and $\mI_c$ into style features $\vf_S$ and content feature $\vf_c$, respectively. \textbf{(b)} Then the cross-attention module is used to learn patch-level attention $\mA_{\textit{patch}}$ between the patches of content and references. \textbf{(c)} The structure-level style enhancement module further utilizes the structure information of Chinese characters to reweight $\mA_{\textit{patch}}$, in order to better learn the structure-level styles. \textbf{(d)} The Transformer module models the font images and predicts the target glyph indices $\vs_{cs}$. \textbf{(e)} The obtained codebook indices $\vs_{cs}$ is used to retrieve quantized token vectors from the font codebook, which is subsequently taken into the font decoder to generate the final image.}}
\label{vqfont_arc}
\end{figure*}

\section{Related Work} 
\subsection{Many-shot Font Generation}
Early methods~\cite{tian2016rewrite,tian2017master,jiang2017dcfont,lyu2017auto,chang2018chinese,sun2018pyramid,jiang2019scfont,yang2019tet,yang2019controllable,gao2020gan,wu2020calligan,wen2021handwritten,hassan2021unpaired} utilize Image-to-Image translation networks to achieve font generation by learning the mapping function between different fonts. 
%zi2zi~\cite{tian2017master} makes simple modifications to pixel2pixel~\cite{isola2017image} to make it suitable for font generation. 
\lxm{Tian~\etal ~presents zi2zi~\cite{tian2017master} which modifies pixel2pixel~\cite{isola2017image} } to make it suitable for font generation.
AGEN~\cite{lyu2017auto} proposes a model for synthesizing Chinese calligraphy images with specified style from standard font images. HGAN~\cite{chang2018chinese} proposes a Hierarchical Generative Adversarial Network consisting of a transfer network and hierarchical adversarial discriminator based on zi2zi. PEGAN~\cite{sun2018pyramid} employs cascaded refinement connections and mirror skip connections to embed a multiscale pyramid of down-sampled input into the encoder feature maps. 
%However, these methods can only transfer glyphs from one domain to another and if there is a need to generate new fonts, the model needs to be retrained.
\lxm{However, these methods can only transfer glyphs from one known domain to another one that has appeared in the training process, making them incapable of generalizing to new fonts.}
% However, these methods can only generate fonts seen during the training phase and is not suitable for unseen fonts.

% \subsection{2.1\quad Image-to-image translation}

% Image-to-image (I2I) translation refers to transferring the style of an image from its source domain to the target domain while preserving the content of the image. Pixel2Pixel ~\cite{isola2017image}, as a representative of GAN-based I2I translation methods ~\cite{shrivastava2017learning,zhu2017unpaired,wang2018high,choi2018stargan}, learns the mapping function from the source domain to the target domain. However, GAN-based methods can only transfer images to a specific domain. To address this limitation, few-shot I2I translation methods ~\cite{benaim2017one,huang2017arbitrary,kim2017learning,chen2018gated,liu2019few,gu2021lofgan,baek2021rethinking} achieve more flexible style transfer by fusing the content representation of content images with the style representation of reference images. In these methods, the transferred style is typically manifested as texture, painting style and color tone. As a special case of I2I translation, font generation usually represents the style as edge details, brush thickness, stroke characteristics and so on.

\subsection{Few-shot Font Generation}
%Few-shot font generation is a task aimed at obtaining a new font library by utilizing a small set of reference glyphs to transfer a standard font library onto the target style. 
\lxm{In comparison, few-shot font generation is more flexible, as it can obtain a new font library by utilizing a few reference glyphs. 
}
Current methods~\cite{sun2017learning,zhang2018separating,gao2019artistic,cha2020few,park2021few,park2021multiple,xie2021dg,liu2022xmp,wang2023cf} disentangle font images into content features and style features to achieve few-shot font generation. SA-VAE~\cite{sun2017learning}, EMD~\cite{zhang2018separating}, and AGISNet~\cite{gao2019artistic} learn global feature representation on font images, but they neglect the design criterion of characters, thereby easily resulting in local details missing. DM-Font~\cite{cha2020few}, LF-Font~\cite{park2021few} and MX-Font~\cite{park2021multiple} explicitly decompose characters into components and learn component-wise feature representation to facilitate the learning of local details. XMP-Font~\cite{liu2022xmp} proposes a self-supervised cross-modality pre-training strategy and a cross-modality transformer-based encoder to model style representation of all scales. DG-Font~\cite{xie2021dg} introduces a feature deformation skip connection to predict displacement maps from the content glyph to the target glyph and its improved version, CF-Font~\cite{wang2023cf}, expands the variety of content fonts and fuses multiple content features by CFM module. Besides, NTF~\cite{fu2023neural} achieves few-shot font generation by modeling it as a continuous transformation process using a neural transformation field. 
%Nevertheless, regardless of whether global or component-wise disentanglement is used, an average operation is typically performed on the extracted features. Therefore, these methods result in loss of fine-grained details. Fs-Font~\cite{tang2022few} no longer explicitly disentangles content features and style features of font images. Instead, it utilizes SAM module based on cross-attention mechanism to match the spatial correspondence between content glyphs and reference glyphs and then achieves font generation by aggregating fine-grained styles.
\lxm{Nevertheless, regardless of whether using global or component-wise disentanglement, an average operation is usually performed on the extracted features, which easily weakens the local information and results in the loss of fine-grained details. FS-Font~\cite{tang2022few} begins to avoid explicitly disentangling content and style features of font images. It utilizes a cross-attention mechanism to match the patch-level correspondence between content and reference glyphs, and then aggregate fine-grained styles for font generation. In this work, we mainly follow the settings of FS-Font and attempt to address the issues of missing details and distorted strokes.}

%\subsection{Vector-Quantized codebook for encoding priors}
\subsection{\lxmup{Codebook for Encapsulating Token Prior}}
\label{Codebook learning}
VQVAE~\cite{van2017neural} is an extension of the autoencoder~\cite{hinton1993autoencoders} that introduces the vector-quantized codebook for the first time. By converting the continuous features into discrete features within a limited space, it resolves the issue of ``posterior collapse" in the autoencoder architecture. To achieve better self-reconstruction results, VQVAE2~\cite{razavi2019generating} introduces a multi-scale codebook. VQGAN~\cite{esser2021taming} enhances generation capabilities and further compresses the codebook size by introducing adversarial loss and perceptual loss. During the training phase, the codebook is continuously updated, thus encapsulating rich priors. Currently, the codebook has been used for many image restoration tasks to recover image details. RIDCP~\cite{wu2023ridcp} utilizes a codebook that stores scene graphs without fog/rain information to achieve image dehazing, FeMaSR~\cite{chen2022real} utilizes a high-resolution prior codebook to achieve image super-resolution, Codeformer~\cite{zhou2022towards} and VQFR~\cite{gu2022vqfr} utilize a codebook that stores high-quality facial textures for blind face restoration. 
MARCONet~\cite{li2023marconet} combines the codebook and StyleGAN~\cite{karras2020analyzing} to generate the specific characters for providing detailed reference in text image super-resolution tasks.
With the aid of the codebook which possesses rich priors, degraded images can be well restored to photo-realistic results. Currently, few-shot font generation tasks also suffer from stroke distortion, detail loss, and other related issues. Inspired by the benefits brought by VQGAN on the aforementioned methods, we introduce a font codebook with rich stroke priors for the first time. 
\lxm{By using Transformer~\cite{vaswani2017attention} to match corresponding code indices, \lxmup{the re-arranged tokens from the codebook} can generate font images with clear strokes and realistic details.}

\section{Method}
%%%%%%%%%
% Method Outline
% 开头介绍Few-shot Font Generation任务定义，并简单概括每段内容
% 1. Preliminary of VQGAN
% 介绍VQGAN定义，重点写在字体上训练时的不同之处
% 2. VQ-Font
% 简述FS-Font作为Baseline，重点强调codebook带来的细节增强
% 3. Structure-level style reweighter
% 
% 4. Training objective
%%%%%%%%%

\begin{figure}[t]
\centering
\includegraphics[width=0.95\columnwidth]{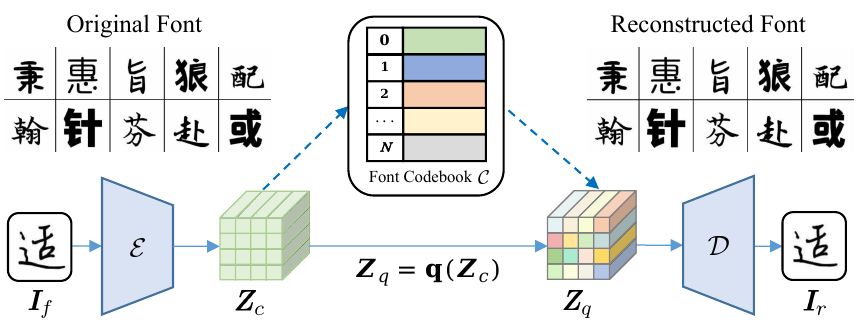} % Reduce the figure size so that it is slightly narrower than the column. Don't use precise values for figure width.This setup will avoid overfull boxes.
\caption{\lxmup{\textbf{The details of our font VQGAN. }The encoder $\mathcal{E}$ first maps the font image $\mI_f$ into continuous feature space $\mZ_c$. Then, $\mZ_c$ is quantized into discrete feature $\mZ_q$ with font codebook $\mathcal{C}$.
% replaced by discrete codes in codebook $\mathcal{C}$ through quantization operation to obtain the discrete feature $\mZ_q$.
Finally, the decoder maps $\mZ_q$ back into the image space to generate result $\mI_r$. Notably, the \textit{out-of-domain} font can be also well reconstructed using the font VQGAN.}
% input and output are shown in the top-left and top-right regions, which do not appear in our training stage.}
} 
\label{vqgan}
\end{figure}

\zyb{
Few-shot font generation transfers a content glyph $\mI_c$ to a new style described by several glyphs $\mI_S=\left\{I_s^i\right\}_{i=1}^k$.
It requires ensuring both the quality of synthesized glyphs and the fidelity of captured styles.
In this section, we introduce a VQGAN-based framework (i.e., VQ-Font) to improve them through token prior refinement and structure-level style enhancement.
Firstly, VQ-Font encapsulates stroke priors by VQGAN-based self-reconstruction (Sec.~\ref{sec:3.1}), and then refines the synthesized strokes with the encapsulated priors (Sec.~\ref{sec:3.2}).
Secondly, it enhances the fine-grained styles using the inherent structure of Chinese characters (Sec.~\ref{sec:3.3}).
The overall training objectives are summarized in Sec.~\ref{sec:3.4}.
}
%%%VQGAN Training loss可以放到附加材料%%%
% \subsection{Pretraining font codebook for font detail priors}
\subsection{Self-Reconstruction for Font Token Prior}
\label{sec:3.1}
% In order to obtain the font codebook with rich priors of font details and the corresponding font decoder, we pre-train VQGAN via self-reconstruction. As shown in~\cref{vqgan}, VQGAN mainly consists of three parts: an encoder $E$, a decoder $D$, and a codebook $\mathcal{C}=\left\{c_k \in \mathbb{R}^d\right\}_{k=1}^K$, where the codebook contains $K$ discrete codes. For a single-channel font image $I_f \in \mathbb{R}^{H \times W \times 1}$, the encoder $E$ first maps $I_f$ to a continuous feature $Z_c \in \mathbb{R}^{h \times w \times d}$, where $d$ represents the dimension of latent codes. Then, each ``pixel'' $Z_c^{(i, j)}$ in $Z_c$ is replaced by the nearest discrete code $c_k$ in the codebook $\mathcal{C}$ to obtain the quantized feature $Z_q$:

\if 0
\zyb{
To encapsulate the high-fidelity token prior, we pre-train a VQGAN by self-reconstructing font images with diverse styles and high quality.
%
As shown in~\cref{vqgan}, VQGAN consists of the encoder $\mathcal{E}$, the decoder $\mathcal{D}$, and a learnable codebook $\mathcal{C}=\left\{c_k \in \mathbb{R}^d\right\}_{k=1}^K$.
%
During self-reconstruction, $\mathcal{E}$ encodes a glyph image $\mI_f \in \mathbb{R}^{H \times W \times 1}$ into continuous features $\mZ_c \in \mathbb{R}^{h \times w \times d}$, while $\mathcal{D}$ projects discrete features $\mZ_q$ into a font image $\mI_r$.
%
$\mZ_q$ is quantized from $\mZ_c$ with font codebook $\mathcal{C}$:
}
\fi

\lxmup{
To learn the token prior and incorporate it into the synthesized fonts, we pre-train a VQGAN by self-reconstructing font images with diverse styles and high quality.
As shown in~\cref{vqgan}, VQGAN consists of an encoder $\mathcal{E}$, a learnable codebook $\mathcal{C}=\left\{c_k \in \mathbb{R}^{d}\right\}_{k=1}^K$, and a decoder $\mathcal{D}$.
During self-reconstruction, $\mathcal{E}$ encodes a glyph image $\mI_f \in \mathbb{R}^{H \times W \times 1}$ into continuous features $\mZ_c \in \mathbb{R}^{h \times w \times d}$.
Then, an element-wise quantization $\mathbf{q}(\cdot)$ is performed to replace each code in $\mZ_c$ with its closest entry in codebook $\mathcal{C}$: 
%while $\mathcal{D}$ projects discrete features $\mZ_q$ into a font image $\mI_r$.
%
%$\mZ_q$ is quantized from $\mZ_c$ with font codebook $\mathcal{C}$:
}
\begin{equation}
\!\mZ_q\!=\!\mathbf{q}(\mZ_c)\!:=\!\left(\underset{c_k \in \mathcal{C}}{\arg \min}\left\|\\\mZ_c^{(i, j)}-c_k\right\|\right)\!\in\! \mathbb{R}^{h \times w \times d}\,.
\label{eq:quantize}
\end{equation}
\lxmup{
The final reconstruction result is obtained through:
\begin{equation}
    % \mI_r = \mathcal{D}\left(\mZ_q\right)\,.
    \mI_r=\mathcal{D}(\mZ_q)=\mathcal{D}(\mathbf{q}(\mathcal{E}(\mI_f))\approx \mI_f\,. 
\end{equation}
}
%where $\mathbf{q}(\cdot)$ represents the operation of quantization. 
%
% Furthermore, we can also obtain a sequence $s\in\{0,\ldots,|{\mathcal{C}}|-1\}^{hw}$ that represents the indices of the replaced codes in codebook:

\zyb{
The indices sequence $s\in\{0,\ldots,|{\mathcal{C}}|-1\}^{hw}$ of $\mZ_c$ in codebook $\mathcal{C}$ is defined as:
}
\begin{equation}
\label{eqn:indice}
s^{(i,j)}=k \quad\text{ such that }\quad \mZ_q^{(i, j)}=c_k\,.
\end{equation} 

% Finally, the decoder $\mathcal{D}$ projects $Z_q$ into a font image $I_r$. 
%

\lxmup{During pre-training, the above three modules (encoder~$\mathcal{E}$, codebook~$\mathcal{C}$, and decoder $\mathcal{D}$) can be optimized in an end-to-end manner.
We follow VQGAN and adopt L1 loss $\mathcal{L}_1$, perceptual loss $\mathcal{L}_{\textit{per}}$~\cite{johnson2016perceptual,zhang2018unreasonable}, and adversarial loss $\mathcal{L}_{\textit{adv}}$~\cite{esser2021taming} between the reconstructed image $\mI_r$ and the input $\mI_f$.
$\mathcal{L}_{\textit{code}}$ and commitment loss $\mathcal{L}_{\textit{comm}}$ are used to update the codebook~$\mathcal{C}$ and encoder $\mathcal{E}$, respectively:
\begin{equation}
\label{eqn:vgg}
\begin{aligned}
\mathcal{L}_1 &=\|\mI_f-\mI_r\|_1\,, \\
\mathcal{L}_{\textit{per}}&=\|\Phi(\mI_{f})-\Phi(\mI_{r})\|_2^2\,,\\
% \mathcal{L}_{\textit{adv}}^D &= [\log \mathfrak{D}(\mI_f)+\log(1-\mathfrak{D}(\mI_r))]\,,\\
\mathcal{L}_{\textit{adv}} &= -\log \mathfrak{D}(\mI_r)\,,\\
\mathcal{L}_{\textit{code}} &=\|\mathbf{sg}(\mZ_c)-\mZ_q\|_2^2\,, \\
\mathcal{L}_{\textit{comm}} &=\|\mZ_c-\mathbf{sg}(\mZ_q)\|_2^2\,,
\end{aligned}
\end{equation}
where $\mathbf{sg}(\cdot)$ indicates the stop gradient operation. $\Phi$ denotes a pre-trained VGG16 model~\cite{simonyan2014very} and $\mathfrak{D}$ represents the discriminator.
}

The overall training loss of VQGAN is summarized as:
\begin{equation}
\mathcal{L}=\mathcal{L}_{1}+\mathcal{L}_{\textit{per}}+\lambda_{\textit{adv}}\mathcal{L}_{\textit{adv}}+\mathcal{L}_{\textit{code}}+\lambda_{\textit{comm}}\mathcal{L}_{\textit{comm}}.
\end{equation}
%In the experiment, we set $\lambda_{comm}$ to $0.5$ and $\lambda_{\textit{adv}}$ to $0.8$.
$\lambda_{\textit{comm}}$ and $\lambda_{\textit{\textit{adv}}}$ are the trade-off parameters and we set them to $0.5$ and $0.8$ in our experiment, respectively.
\if 0
\begin{equation}
\begin{split}
\mathcal{L}&=\mathcal{L}_{1}+\mathcal{L}_{\textit{per}}+\lambda_{\textit{adv}}\mathcal{L}_{\textit{adv}} \\
&+\mathcal{L}_{\textit{code}}+\lambda_{\textit{comm}}\mathcal{L}_{\textit{comm}}.
\end{split}
\end{equation}
\fi

% The above self-reconstruction process is summarized as:\\
% \begin{equation}
% I_r=D(Z_q)=D(\mathbf{q}(E(I_f))\approx I_f. 
% \end{equation}

\if 0
\zyb{
Notably, despite diverse font styles and character structures, the well-trained VQGAN demonstrates nearly perfect reconstruction on the out-of-domain glyphs.
As shown in~\cref{vqgan}, almost all stroke details are well recovered with the quantization and decoding processes, showing high-fidelity stroke priors in the codebook and decoder.
}
\fi

\lxmup{
Notably, our VQGAN demonstrates remarkable generalization capabilities across out-of-domain glyphs and styles.
As shown in~\cref{vqgan}, although these font styles and characters do not appear in the training process, nearly all stroke details are well reconstructed with the quantization process, showing the ability in generalizing to different font images.
}
\subsection{Token Prior Refinement}
\label{sec:3.2}

{
To effectively integrate the token prior, VQ-Font leverages the well-trained codebook $\mathcal{C}$ and decoder $\mathcal{D}$. It casts font generation task into indices prediction task.
This process involves aggregating fine-grained styles from reference glyphs and predicting codebook indices of ground-truth glyph $\mI_g$.
}
{
\paragraph{Styles aggregation.}
Following FS-Font, our VQ-Font employs a cross-attention module to attentively capture fine-grained styles, where it takes a content glyph $\mI_c$ as query, and $k$ reference glyphs $\mI_S$ as key and value.
Specifically, we first use a content encoder $\mathcal{E}_c$ and a style encoder $\mathcal{E}_s$ to extract the feature maps from $\mI_c$ and $\mI_S$, \ie, $\vf_c \in\mathbb{R}^{hw\times c}$ and $\vf_S \in\mathbb{R}^{khw\times c}$.
Then, we calculate their attention weights as: 
\begin{equation}
    \mA_{\textit{patch}} = \frac{(\mW^Q \vf_c)(\mW^K \vf_S)^T} {\sqrt{c}}\,,
\end{equation}
where $\mW^Q$ and $\mW^K$ are learnable parameters to project the extracted features into query and key, respectively. $c$ is set to 256.
With the above weights, VQ-Font attentively captures patch-level styles from reference features $\vf_S$ to obtain the aggregated features $\vf_{cs}$, which is formulated as:
{
\begin{equation}
    \vf_{cs}= \mathrm{Softmax}(\mA_{\textit{patch}}) \cdot (\mW^V \vf_S)\,,
    \label{eq:full_attn}
\end{equation}
}
%$\mW^V$ is learnable parameter.
}
\lxmup{In this way, we obtain patch-level aggregation features from the references, which may easily generate distorted strokes. Therefore, we need further fine-tuning for better quality.}
\zyb{
\paragraph{Vector-Quantized font generation.}
To exploit the token prior in VQGAN, VQ-Font aims to quantize $\vf_{cs}$ into $\vf_{q}$ according to font codebook $\mathcal{C}$.
However, due to the discrepancy between the feature spaces of $\vf_{cs}$ and the VQGAN encoded, it is infeasible to compute the indices sequence of $\vf_{cs}$ by nearest neighbor lookup.
Inspired by~\cite{zhou2022towards}, we utilize a Transformer module to predict the indices $\vs_{cs}\in\{0,\ldots,|\mathcal{C}|-1\}^{hw}$ for all patch tokens of $\vf_{cs}$.
It employs $15$ self-attention layers to globally model all tokens and an MLP to classify each token.
Given the target glyph $\mI_g$, we optimize this module from two aspects:
(i) index prediction and (ii) image regression.
%
%Intuitively, the indices of $\vf_{cs}$ can be approximated by encoding and quantizing $\mI_g$, and serves as the ground-truth of prediction.
Intuitively, the indices of $\vf_{cs}$ can be approximated to those indices obtained by encoding and quantizing ground-truth font $\mI_g$.
Moreover, a glyph image $\mI_{q}$ is projected from $\vf_q$ with the VQGAN decoder and is combined with $\mI_g$ to obtain reconstruction loss.
}
\lxmup{To generalize the token prior to this task and preserve the effectiveness of prior knowledge, we fix the font codebook and only fine-tune the former layers of the decoder.}
%Notably, to ensure the effectiveness of the prior knowledge, we fix the font codebook and only fine-tune the former layers of decoder. 
% we prevent the gradient backpropagating to the transformer module and the previous modules. Thus, during each iteration, the earlier part of the model focuses on index prediction, while the later part of the model focuses on font generation.

\begin{figure}[t]
\centering
\includegraphics[width=0.9\columnwidth]{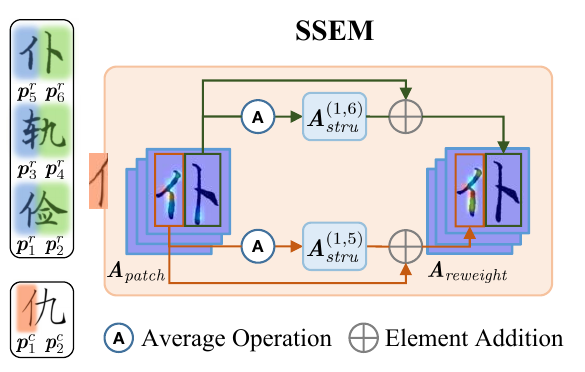} % Reduce the figure size so that it is slightly narrower than the column. Don't use precise values for figure width.This setup will avoid overfull boxes.
\caption{\textbf{The overview of our Structure-level Style Enhancement Module~(SSEM).}}
\label{ssrm}
\end{figure}

% \subsection{Structure-level style reweighting module}
\subsection{Structure-level Style Enhancement (SSEM)}
\label{sec:3.3}

 {
Although leveraging the learned priors could promisingly improve the quality of synthesized glyphs, there are still inconsistent fine-grained styles between synthesized and reference glyphs, \eg, the last second row in Fig.~\ref{ablation_1}.
Compared to style transfer in RGB images, few-shot font generation in single-channel glyphs is more likely to acquire irrelevant patch-level styles.
The main reason is that the original attention weights $\mA_{\textit{patch}}$ are mainly based on geometry.
To enhance the fidelity of captured styles, we propose to utilize the inherent structure (See Fig.~\ref{12_stru}) of Chinese characters to further recalibrate the attention weights.
}

Based on~\cref{12_stru}, we decompose the content glyph and reference glyphs into the structure components $\{\vp_i^c\}_{i=0}^{m}$ and $\{\vp_j^r\}_{j=0}^{n}$, where $\vp_i^c$ and $\vp_j^r$ denote the set of patch positions.
After that, we calculate the structure-level attention weights $\mA_{\textit{stru}} \in \mathbb{R}^{mn}$.
%
% We first sum up $\mA_{patch}$ along head-dimension to obtain the global attention weights $\mA_{patch}^{global}\in\mathbb{R}^{hw\times khw}$. 
%
% As illustrated in~\cref{ssrm}, we locate the positions in the content dimension $(hw)_{ci}$ and style dimension $(khw)_{si}$ of $\mA_{patch}$ that correspond to the different structure components of the content and reference glyphs, respectively. 
%
As illustrated in~\cref{ssrm}, the attention weight $\mA_{\textit{stru}}^{(i,j)}$ between $\vp_i^c$ and $\vp_j^r$ is obtained by averaging their corresponding patch-level weights:
\begin{equation} 
\mA_{\textit{stru}}^{(i,j)} =\frac{1}{|\vp_i^c|\cdot|\vp_j^r|} \sum_{x\in \vp_i^c, y\in \vp_j^r} \mA_{\textit{patch}}^{(x,y)}.
\end{equation}
% where $m$ and $n$ represent the number of structure components in the content and reference glyphs, respectively,and in this example, $m$=2, $n$=6, $(hw)_{ci}\in\left\{c_1,c_2\right\}$, $(khw)_{si}\in\left\{s_1,s_2,s_3,s_4,s_5,s_6\right\}$. 
Finally, we reweight $\mA_{\textit{patch}}$ by adding $\mA_{\textit{stru}}$ to the corresponding patch positions:
\begin{equation} 
\mA_{\textit{reweight}} = \mA_{\textit{patch}}\oplus \mA_{\textit{stru}}\,.
\end{equation}
%\lxmup{After that, we obtain new $\vf_{cs}$ concentrated on both patch-level and structure-level.}
\lxmup{The new fusion feature $\vf_{cs}$ in~\cref{eq:full_attn} is reformulated as:}
\begin{equation}
    \vf^*_{cs}= \mathrm{Softmax}(\mA_{\textit{\textit{reweight}}}) \cdot (\mW^V \vf_S)\,.
    \label{eq:stru_attn}
\end{equation}
% Finally, we multiply $\hat{V}$ and A together to obtain the fused feature $f_{atten}$:
% \begin{equation} 
% f_{atten} = softmax(A)\hat{V}^{\top}\in\mathbb{R}^{H \times hw\times c_H}
% \end{equation}

\lxmup{In this way, our attention map can concentrate more on matching corresponding structure components, and reduce the adverse effect of other irrelevant strokes. Fig.~\ref{atten_map} shows that after SSEM, our attention map $A_\textit{reweight}$ has higher attention in corresponding structure components, thereby benefiting the following style transformation.}
%In Fig.~\ref{atten_map}, our structure-level style enhancement effectively decreases the attention weights on irrelevant tokens, thereby visibly improving the fidelity in fine-grained styles.

\subsection{Training Objective}
\label{sec:3.4}
%Given a content glyph ${I_c}$ with content c and several reference glyphs ${I_S}$ with style s, our model can generate font images ${I_g}$ with both content c and style s. 

We train the content encoder $\mathcal{E}_c$, style encoder $\mathcal{E}_s$, cross-attention module, and Transformer module using cross-entropy loss $\mathcal{L}_{\textit{indice}}$ while keeping font codebook fixed. We fine-tune the first four layers of the pre-trained font decoder using VQGAN-like losses, including L1 loss $\mathcal{L}_1$, perceptual loss $\mathcal{L}_{\textit{per}}$, and adversarial loss $\mathcal{L}_{\textit{adv}}$.
\subsubsection{Cross Entropy loss.}
%We input the ground truth font image ${\mI_{g}}$ into the pre-trained VQGAN to obtain the ground truth font codebook indices $\vs_{g}$. To further improve the prediction performance of our model, we design a self-reconstruction branch that uses ${\mI_{g}}$ as the reference glyph.
\lxmup{We first obtain the ground-truth codebook indices~$\vs_{g}$ using~\cref{eqn:indice} by taking the ground-truth font image~${\mI_{g}}$ into the pre-trained VQGAN. To further improve the prediction performance, we follow FS-Font and design a self-reconstruction branch that uses ${\mI_{g}}$ as the reference glyph. The code indices learning is defined as:}
\begin{equation}
\mathcal{L}_{\textit{indice}}^{\textit{main}}=\mathrm{CE}(\hat{\vs}_{cs},\vs_{g});\quad \mathcal{L}_{\textit{indice}}^{\textit{self}}=\mathrm{CE}(\widetilde{\vs}_{cs},\vs_{g}), 
\end{equation}
where $\hat{\vs}_{cs}$ represents the indices predicted by the main branch and $\widetilde{\vs}_{cs}$ represents the indices predicted by the self-reconstruction branch. %This brings to better alignment
\subsubsection{L1 loss.} 
To maintain pixel-level consistency between the generated font images $\mI_{q}$ and the ground-truth font images $\mI_g$, we employ L1 loss as our reconstruction loss:
\begin{equation}
\mathcal{L}_{1}^f=\|\mI_{g}-\mI_{q}||_{1}.
\end{equation}
\subsubsection{Adversarial loss and Perceptual loss.}
 To further ensure that the generated font images have high visual quality, we additionally utilize adversarial loss and perceptual loss. Moreover, in our experiments, we employ a multi-head projection discriminator~\cite{park2021few} and use the Unicode encoding of each Chinese character as the label:
\begin{equation}
\begin{split}
\begin{aligned}
\mathcal{L}_{\textit{adv}}^D = &-\mathbb{E}_{\mI_{g}\sim p_{\textit{data}}}\max{(0,-1+D_{\textit{uni}}(\mI_{g}))} \\
&-\mathbb{E}_{\mI_q\sim p_{\textit{gen}}}\max{(0,-1-D_{\textit{uni}}(\mI_q))}\,, \\
\mathcal{L}_{\textit{adv}}^G=&-\mathbb{E}_{\mI_q\sim p_{\textit{gen}}}D_{\textit{uni}}(\mI_q)\,, \\
\mathcal{L}_{\textit{per}}^f=&\,\,\|\Phi(\mI_{g})-\Phi(\mI_{q})\|_2^2,
\end{aligned}
\end{split}
\end{equation}
where $\Phi$ denotes VGG16 same as that in~\cref{eqn:vgg}.
\subsubsection{Overall objective loss.}
To sum up, the final loss function for training our VQ-Font is formulated as:
\begin{equation}
\begin{split}
\mathcal{L}_{\textit{VQ-Font}} =&\lambda_{\textit{self}}\mathcal{L}_{\textit{indice}}^{\textit{self}}+\lambda_{\textit{main}}\mathcal{L}_{\textit{indice}}^{\textit{main}}+\lambda_{1}^f\mathcal{L}_{1}^f \\
+&\lambda_{\textit{adv}}^f\mathcal{L}_{\textit{adv}}^G+\lambda_{\textit{per}}^f\mathcal{L}_{\textit{per}}^f\,,
\end{split}
\end{equation}
%$\mathcal{L}_{vqfont}$ is controlled by hyperparameters $\lambda_{\textit{self}}$, $\lambda_{main}$, $\lambda_{1}$, $\lambda_{adv}$ and $\lambda_{per}$, which determine the relative weights of each objective. Throughout all the experiments, we set $\lambda_{\textit{main}}$ = $\lambda_{1}$ = 2, $\lambda_{self}$ = $\lambda_{per}$ = 1 and $\lambda_{adv}$ = 0.002.
\lxmup{where $\lambda_{\textit{self}}$, $\lambda_{\textit{main}}$, $\lambda_{1}^f$, $\lambda_{\textit{adv}}^f$ and $\lambda_{\textit{per}}^f$ are the trade-off parameters for balancing each loss item. In our experiments, we set $\lambda_{\textit{main}}$ = $\lambda_{1}^f$ = 2, $\lambda_{\textit{self}}$ = $\lambda_{\textit{per}}^f$ = 1 and $\lambda_{\textit{adv}}^f$ = 0.002.}

\begin{table*}[!tb]
%\vspace{-2pt}
%\setlength{\belowcaptionskip}{-4pt}
\centering
\begin{tabular} { p{3.3cm}<{\centering} | p{1.3cm}<{\centering} p{1.3cm}<{\centering} p{1.3cm}<{\centering} p{1.3cm}<{\centering}  p{1.3cm}<{\centering} | p{1.8cm}<{\centering}  p{1.8cm}<{\centering} }
%\Xhline{1pt}
\toprule
\multicolumn{8}{c}{\textbf{Seen Fonts Unseen Chars} \textbf{(SFUC)}} \\
\midrule
Method &  L1~$\downarrow$ & RMSE~$\downarrow$ & PSNR~$\uparrow$ & SSIM~$\uparrow$ & LPIPS~$\downarrow$ & User (C)\%~$\uparrow$ & User (S)\%~$\uparrow$  \\
\midrule
LF-Font (AAAI 2021) &0.0921 &0.264  &17.818 &0.746 &0.157   &70.5   &4.0  \\
MX-Font (ICCV 2021) &0.1002 &0.278 &17.326   & 0.725   &0.169   &83.4   &2.9   \\
DG-Font (CVPR 2021) &0.0747 &0.233 &18.901 &0.782 &0.127 &80.7   &6.7   \\
FS-Font (CVPR 2022) &\underline{0.0663} & 0.220 &\underline{19.702} &\underline{0.805} &0.126 &\underline{90.3}   &\underline{12.9}   \\
CF-Font (CVPR 2023) &0.0667 &\underline{0.217} &19.559 &\underline{0.805} &\underline{0.111} &86.7   &10.6   \\
VQ-Font (Ours) &\textbf{0.0610} &\textbf{0.209} &\textbf{20.285} &\textbf{0.822} &\textbf{0.096} &\textbf{97.2}   &\textbf{62.9}      \\
\toprule
\multicolumn{8}{c}{\textbf{Unseen Fonts Unseen Chars} \textbf{(UFUC)}} \\

\midrule
Method &  L1~$\downarrow$ & RMSE~$\downarrow$ & PSNR~$\uparrow$ & SSIM~$\uparrow$ & LPIPS~$\downarrow$ & User (C)\%~$\uparrow$ & User (S)\%~$\uparrow$  \\
\midrule

LF-Font (AAAI 2021) &0.0976   &0.274   &17.495   &0.726   &0.174   &68.0   &3.4   \\
MX-Font (ICCV 2021) &0.1061   &0.288   &17.045   &0.706   &0.185   &76.5   &4.0   \\
DG-Font (CVPR 2021) &0.0807   &0.246   &18.465   &0.768   &0.139   &78.4   &4.7   \\
FS-Font (CVPR 2022) &\underline{0.0666}   &\underline{0.220}  &\underline{19.672}   & 0.797 &0.137   &\underline{84.5}   &9.2   \\
CF-Font (CVPR 2023) &0.0685   &0.222   &19.344   &\underline{0.798}   &\underline{0.116}   &84.3   &\underline{11.3}   \\
VQ-Font (Ours)  &\textbf{0.0621} &\textbf{0.210}  &\textbf{20.249}   &\textbf{0.812}  &\textbf{0.103}   &\textbf{95.6}   &\textbf{67.4}   \\
\toprule
\end{tabular}
\caption{\textbf{Quantitative comparison with state-of-the-art methods on SFUC and UFUC datasets.}}
\label{tab:Quantitative_comparison}
\end{table*}

\begin{figure*}[!h]
\centering
\includegraphics[width=0.96\textwidth, height=5.58cm]{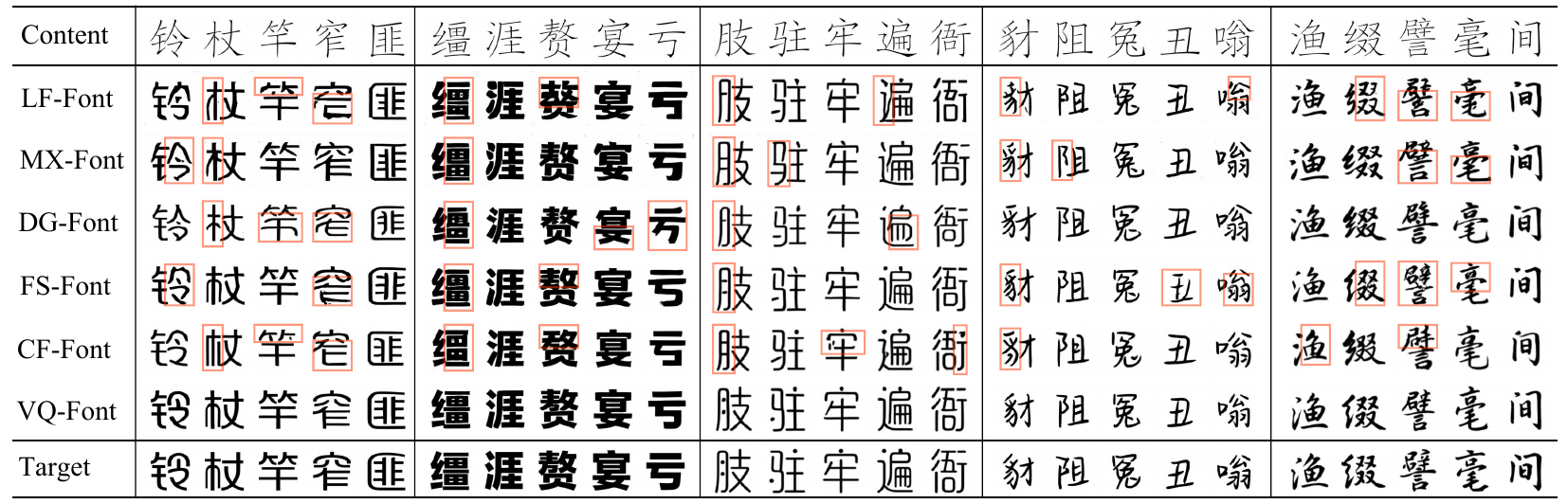} % Reduce the figure size so that it is slightly narrower than the column.
\caption{\textbf{Qualitative comparison with competing methods on UFUC dataset. 
%Results best seen at 500\% zoom.
Best view it by zooming in on the screen.
}}
\label{result}
\end{figure*}

\section{Experiments}
\subsection{Datasets and Evaluation Metrics}
%Due to copyright constraints related to fonts, there is currently no appropriate high-quality publicly font dataset available. As a result, we proceed to create a dataset using the same methodology employed in previous works~\cite{park2021few,park2021multiple,tang2022few}. We collect 382 fonts of various types from the internet to form our dataset, with each font containing 3499 Chinese characters at a resolution of 128*128. 
\lxm{
%Since there is no publicly available font dataset for this task, 
We follow previous works~\cite{park2021few,park2021multiple,tang2022few} and collect 382 fonts with various types to build our dataset. Each font contains 3499 Chinese characters and the resolution for each character is $128\times128$.
We split these 3499 characters into 3 groups, \ie, 2841 seen characters, 158 reference characters, and 500 unseen characters.
For each character, we follow FS-Font~\cite{tang2022few} and select 3 reference characters from the reference set that can cover most of its structure components. We use Kai font as our default content font and train our model on 371 seen fonts, leaving 10 unseen fonts that do not appear in the training stage. In this way, our training set totally consists of 371 seen fonts, each of which has 2841 seen characters (SFSC). Our test set consists of two parts, \ie, 10 seen fonts with 500 unseen characters (SFUC) and 10 unseen fonts with 500 unseen characters (UFUC), encompassing a diverse range of font types, such as handwriting, printing, and artistic styles.}

%To assess the quality of the generated font images, we compare our method with other state-of-the-art methods using several metrics that focus on pixel consistency, including L1, RMSE, PSNR, and SSIM, as well as the perceptual relevance metric LPIPS~\cite{zhang2018unreasonable}. Additionally, we conduct a user study to further demonstrate the superiority of our method, which is described in detail in~\cref{Comparison_methods}.

\lxm{To evaluate the performance, we follow these competing methods and report the L1, RMSE, PSNR, SSIM and LPIPS~\cite{zhang2018unreasonable}, covering both pixel consistency and perceptual similarity. Additionally, we also conduct a user study to further assess the visual quality of the generated results from human perception.}

%%%这一节可以放到附加材料%%%
\subsection{Implementation Details}
%In the process of pre-training the font codebook and font decoder, we encode 128x128x1 font images into 16x16 feature maps and set the size of the font codebook to 1024. At this stage, VQGAN is trained for 200k iterations with a learning rate of 0.000045. During the training of VQ-Font, we set the number of attention heads in cross attention module to 8 and select 3 reference characters for each Chinese character. We keep the font codebook and part of the font decoder fixed, and train VQ-Font for 300k iterations with a learning rate of 0.0002 for the discriminator and 0.0008 for the generator. For all stages of training, we use Adam optimizer~\cite{kingma2014adam} with a batch size of 32, and train the model using a NVIDIA A6000 GPU on the PyTorch framework.

\lxm{In the pre-training phase of our font codebook, we encode the font images into $16\times16$ features. The size of our font codebook is set to 1,024. At this stage, VQGAN is trained for $2e5$ iterations with a learning rate of $4e\textit{-}5$. The number of attention heads in the cross-attention module is set to 8. We select 3 reference characters for each Chinese character. In the subsequent token prior refinement stage, we keep the pre-trained font codebook and the later layers of the decoder fixed, while concentrating on training the remaining layers of the VQ-Font for $300$k iterations. Here, the learning rate is set to $2e\textit{-}4$. We adopt Adam optimizer~\cite{kingma2014adam} with a batch size of 32 and rely on one A6000 GPU.}

\subsection{Comparison Methods}
\label{Comparison_methods}
We compare the performance of our proposed VQ-Font with five state-of-the-art methods, including LF-Font~\cite{park2021few}, MX-Font~\cite{park2021multiple}, DG-Font~\cite{xie2021dg}, FS-Font~\cite{tang2022few}, and CF-Font~\cite{wang2023cf}. 
%To ensure the fairness of the experimental results, we retrain all these methods on our dataset and select Kai font as the content font.
\lxm{To make a fair comparison, we retrain all these methods using their default settings on our dataset.
More results and analyses can be found in our suppl.}

\subsubsection{Quantitative evaluation.}
\cref{tab:Quantitative_comparison} presents the comparison of our VQ-Font with other state-of-the-art methods. The results demonstrate that our approach outperforms others in terms of both pixel-based and perception-based metrics on UFUC and SFUC datasets. %Specifically, we achieve 13.51\% improvement over the second-best result on SFUC dataset and 11.21\% improvement on UFUC dataset in terms of LPIPS, 2.93\% improvement on SFUC dataset and 2.96\% on UFUC dataset in terms of PSNR.
\lxm{Specifically, in terms of L1, we obtain 7.99\% improvement over the second-best result on SFUC dataset and 6.76\% on UFUC dataset, respectively.
As for LPIPS, our VQ-Font outperforms the second-best result with a large margin, \ie, 13.51\% improvement on SFUC dataset and 11.21\% improvement on UFUC dataset.
This indicates that the synthesized results of our VQ-Font have a better fidelity and also align better with human perception.
}

\vspace{-5pt}
\subsubsection{Qualitative evaluation.}
%As shown in~\cref{result}, we select various styles of fonts, including serif, artistic, and handwritten fonts, from UFUC dataset for qualitative comparison. We can observe that LF-font and MX-font produce stroke artifacts and fail to capture the fine-grained styles. DG-Font and CF-Font exhibit missing strokes and distorted strokes in some challenging cases. Fs-Font tends to produce font blurring and loss of stroke details. Our proposed VQ-Font, on the other hand, can accurately preserve the content of the character while effectively learning the stroke-level and structure-level styles of the reference glyphs.
\lxm{As shown in~\cref{result}, we select various styles of fonts, including serif, artistic, and handwriting fonts, from UFUC dataset for qualitative comparison. We can observe that LF-font and MX-font struggle to capture the fine-grained styles, resulting in stroke artifacts (see \textcolor{red}{red boxes}). DG-Font and CF-Font also exhibit missing strokes and distorted strokes in challenging cases (\eg, 2$\sim$3 columns in~\cref{result}). Besides, FS-Font tends to produce blurry font and lose stroke details (\eg, 5$\sim$6 columns in~\cref{result}). In comparison, our proposed VQ-Font can effectively capture and transfer the stroke-level and structure-level styles of the reference glyphs. The integration of token prior further contributes to the higher quality of the glyphs.}

\subsubsection{User study.}
%To further compare the visual quality of different methods, we conduct a user study. We recruit a total of 30 volunteers to rate the experimental results. We first select 10 fonts from both UFUC and SFUC datasets respectively, and then randomly generate 10 characters for each font using above methods. For content accuracy, we ask the volunteers to select font images with clear content performance, free from extra strokes or other anomalies from the results generated by each method. For style consistency, we instruct the volunteers to find results from different methods that are closest to the reference font, with consistent stroke features, stroke thickness and edge details.
%The final result shows that VQ-Font surpasses other methods in both content accuracy and style consistency, thus demonstrating that our method effectively captures the styles from reference glyphs while preserve the content of the character.

\lxm{To further compare the visual quality of different methods, we conduct a user study. A total of 30 volunteers with computer vision backgrounds participated in evaluating the experimental results. We {utilize} 10 fonts with challenging styles (\eg, 2$\sim$4 columns in~\cref{result}) in UFUC and SFUC datasets respectively, and randomly generate 10 characters for each font using these methods. Here we consider two items, \ie, 1) content accuracy which classifies whether the font images within each method have extra strokes and other anomalies or not, and 2) style consistency which concentrates on selecting the font images among all the methods that are closest to the reference in stroke style, stroke thickness, and edge details.
The evaluation result on the right part of~\cref{tab:Quantitative_comparison} demonstrates that 
%on the one hand, nearly all the users think our method has better content accuracy than others. 
on the one hand, our method has better content accuracy than others from human perception.
On the other hand, users are more likely to select our results which have better style consistency, while other methods struggle with these challenging font styles.
%especially in preserving the styles from the reference in such challenging styles.
 }

\begin{table}[t]
\centering
%\vspace{1mm}
\begin{tabular}{ p{1.15cm} | p{0.86cm}<{\centering}  p{0.9cm}<{\centering} p{0.9cm}<{\centering} p{0.9cm}<{\centering} p{0.9cm}<{\centering}}
\toprule
  &  L1~$\downarrow$  & RMSE~$\downarrow$ & PSNR~$\uparrow$ & SSIM~$\uparrow$ & LPIPS~$\downarrow$ \\
\midrule
Baseline &0.0678   &0.223  &19.318  &0.793  &0.116\\
\midrule
+C &0.0628  &0.212  &20.170  &0.810  &0.105 \\
+CS  &\textbf{0.0621} &\textbf{0.210}  &\textbf{20.249}   &\textbf{0.812}  &\textbf{0.103}  \\
\bottomrule
\end{tabular}
%\caption{\textbf{Quantitative analysis of the effectiveness of the font codebook and SSEM.} C is the abbreviation for font codebook, and S for SSEM.}
\caption{\textbf{Quantitative results of different VQ-Font variants.} C and S represent codebook and SSEM, respectively.}
\label{ab_stru_coebook}
\end{table}

\begin{figure}[t]
\centering
\includegraphics[width=0.98\columnwidth]{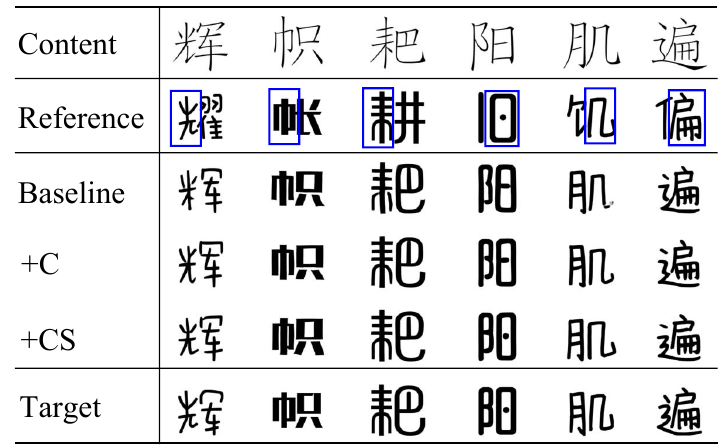} % Reduce the figure size so that it is slightly narrower than the column. Don't use precise values for figure width.This setup will avoid overfull boxes.
%\caption{\textbf{Qualitative results of adding font codebook and SSEM Sequentially.} The structural components appearing in the reference glyphs are highlighted with blue boxes.}
\caption{\lxm{\textbf{Qualitative results of different VQ-Font variants.} The structure components appearing in the reference glyphs are highlighted with \textcolor{blue}{blue boxes}.}}
\label{ablation_1}
\end{figure}

\subsection{Ablation Study}
%In this section, we mainly discuss the effectiveness of font codebook and structure-level style reweighting module. We first train a baseline model by ultilizing the cross-attention mechanism to learn the spatial correspondence at patch-level between content glyphs and reference glyphs. Then, we sequentially add the pre-trained font codebook and structure-level style reweighting module to the baseline model for validation. The ablation results are presented in~\cref{ab_stru_coebook}, after adding the pre-trained font codebook(+C), the metrics are greatly improved. Furthermore, after adding structure-level style reweighting module(+CS), we achieve state-of-the-art results.

\lxm{In this section, we mainly discuss the effectiveness of token prior refinement and structure-aware enhancement. We first train a baseline model by utilizing the cross-attention mechanism to learn the spatial correspondence at the patch level which is similar to FS-Font. Then, we gradually add the pre-trained font codebook and structure-level style enhancement module~(SSEM) to the baseline for validation. \cref{ab_stru_coebook} shows the quantitative results on UFUC. 
One can see that the pre-trained font codebook~(+C) can obviously improve the SSIM and LPIPS performance, which indicates the better visual quality brought by our font codebook. When combining the codebook with the structure-level style enhancement module~(+CS), our method has a further improvement, especially in PSNR. This indicates that SSEM can effectively capture the reference styles and contribute to higher fidelity.}

%The visual results, as depicted in~\cref{ablation_1}, demonstrate a noticeable reduction in distorted strokes and missing details when utilizing the font codebook. Furthermore, the integration of SSRM enhances the fidelity of the generated glyphs by aligning them more accurately with the corresponding structure components in the reference. These improved alignments are highlighted within the blue box, indicating the enhanced alignment achieved by the proposed method.

\begin{figure}[!h]
\setlength{\belowcaptionskip}{-2pt}
\centering
\includegraphics[width=0.96\columnwidth]{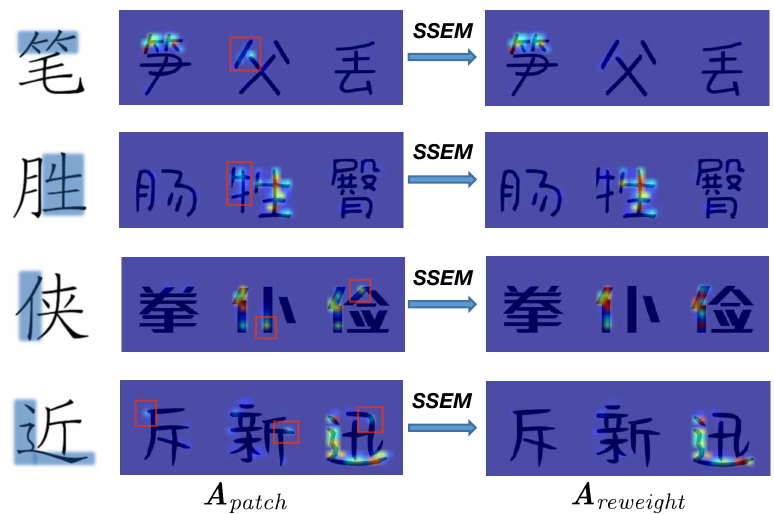} % Reduce the figure size so that it is slightly narrower than the column. Don't use precise values for figure width.This setup will avoid overfull boxes.
\caption{\textbf{Attention maps w/o and w/ SSEM.}}
%\textcolor{red}{Red boxes} highlights the recalibrated parts.
\label{atten_map}
\end{figure}

\begin{table}[h]
\setlength{\belowcaptionskip}{-6pt}
\centering
%\vspace{1mm}
\begin{tabular}{ p{1.36cm}<{\centering} | p{0.85cm}<{\centering}  p{0.88cm}<{\centering} p{0.88cm}<{\centering} p{0.88cm}<{\centering} p{0.88cm}<{\centering}}
\toprule
Decoder & L1~$\downarrow$ & RMSE~$\downarrow$   & PSNR~$\uparrow$   & SSIM~$\uparrow$  & LPIPS~$\downarrow$   \\
\midrule
Fix &0.0648   &0.216  &20.069  &0.805  &0.121\\
Fine-tune &\textbf{0.0621} &\textbf{0.210}  &\textbf{20.249}   &\textbf{0.812}  &\textbf{0.103} \\
\bottomrule
\end{tabular}
%\caption{\textbf{Ablation study on fine-tuning  pretrained font decoder.}}
\caption{\textbf{\lxm{Comparison of fixed and fine-tuned decoders.}}}
\label{decoder}
\end{table}

\lxm{The visual comparison in~\cref{ablation_1} demonstrates a noticeable reduction in distorted strokes and missing details when utilizing the font codebook~(+C). Besides, our SSEM enhances the fidelity of the generated glyphs by aligning them more accurately with the corresponding structure components in the reference. These improved regions are highlighted within the blue box. Both the quantitative and qualitative evaluations demonstrate the improvements in fidelity and visual quality brought by our token prior refinement and structure-aware enhancement.}

\lxm{To further validate the effectiveness of SSEM, we visualize the attention maps before and after enhancement operation. As shown in~\cref{atten_map}, the leftmost column represents the content glyphs, and the visualized components are highlighted in blue. The visualization of different structure components demonstrates that our enhancement operation can eliminate the attention on irrelevant regions (see \textcolor{red}{red boxes}) and  concentrate more on the corresponding components of the reference. This helps to capture the structure-level styles and then boosts the subsequent style transformation.}

%Finally, we explore the effect of fine-tuning the pre-trained font decoder in our method. We conduct experiments by freezing all parameters of the decoder and selectively fine-tuning the first three layers of the decoder. Based on the experimental results in~\cref{decoder}, it is evident that fine-tuning the decoder is more advantageous for style transfer in our method.

\lxm{Finally, we explore the effect of fine-tuning the pre-trained font decoder in our method. We conduct another experiment by freezing all parameters of the decoder. From~\cref{decoder} we can see that although the pre-trained decoder has the ability to generate photo-realistic font images, by fine-tuning the decoder with the end-to-end optimization, it can well generalize to our font generation task and bridge the domain gap between our synthetic and real-world fonts.}

\section{Conclusion}
%In this paper, we propose a VQGAN-based few-shot font generation method named VQ-Font which refines font images by incorporating structure prior encapsulating in font codebook, thus making the generated results more closely resemble real glyphs. Additionally, our proposed Structure-level Style Reweighting Module(SSRM) leverages the structural information of Chinese characters to recalibrate fine-grained styles from the references. This enhances the alignment of structure-level styles between content glyphs and reference glyphs. After structure prior refinement and structure-level style enhancement, we achieve start-of-the art results.

\lxm{In this paper, we propose VQ-Font, a new few-shot font generation framework. It refines the font images by incorporating token prior encapsulated in a pre-trained font codebook. Additionally, the Structure-level Style Enhancement Module~(SSEM) leverages the structure information of Chinese characters to recalibrate fine-grained styles from the references. This enhances the alignment of structure-level styles between content and reference glyphs.
By combining the token prior refinement and SSEM, the results of our VQ-Font are more realistic and have higher fidelity in comparison with other start-of-the-art methods.}

\bibliography{aaai24}

\clearpage

\appendix
 \section{Supplementary Material}
In the supplementary material, we first introduce the architecture of Transformer module used for predicting the glyph indices. Then, we analyze the impact of font codebook size on the generated font images. Next, we describe how we partition Chinese characters into structure components with structure-level style enhancement module. 
Finally, we present additional font generation results and examples of different Chinese character structures.
\subsection{Transformer Module}
\begin{figure}[t]
\centering
\includegraphics[width=0.5\columnwidth]{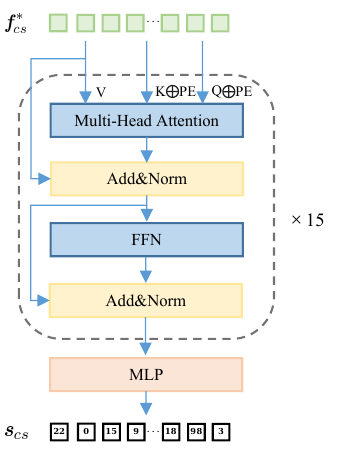} % Reduce the figure size so that it is slightly narrower than the column. Don't use precise values for figure width.This setup will avoid overfull boxes.
\caption{\textbf{The illustration of the Transformer module.}}
\label{Transformer}
\end{figure}

As shown in~\cref{Transformer}, our Transformer module consists of a total of 15 blocks, where each block includes a multi-head self-attention module and a feed-forward network~(FFN). The input for the first block is the fusion feature $\vf^*_{cs}$, while the remaining blocks receive the output of the previous block as their input. We set the number of the attention head to 8 in the multi-head self-attention module. Furthermore, we incorporate learnable positional embeddings~(PE) on Keys and Queries. These positional embeddings provide spatial information and help the model understand the relative positions of different patches within $\vf^*_{cs}$. To obtain the indices sequence $\vs_{cs}$ corresponding to the target glyph, we input $\vf^*_{cs}$ aggregating both patch-level and structure-level styles into the Transformer module and subsequently an MLP~(Multi-Layer Perceptron) is employed to classify each token. By replacing $\vs_{cs}$ with the discrete codes of the codebook, our method can effectively encapsulate the token prior to the synthesized glyphs.

\subsection{Impact of Font Codebook Size}

In our model, the font codebook plays a crucial role that makes the synthesized glyphs more closely to the real-world glyphs and reduces the occurrence of distorted strokes. Inspired by previous works~\cite{chen2022real,zhou2022towards,gu2022vqfr,li2023marconet,wu2023ridcp}, the size of the codebook affects the overall expressive capability. To investigate this in our model, we conduct an ablation study on the size of the font codebook. During the experiments, we set the codebook size to 128, 512, and 1024, while keeping other model settings consistent. We employ L1, RMSE, PSNR, SSIM, and LPIPS~\cite{zhang2018unreasonable} as metrics to quantify the experimental results on UFUC dataset. With the increase of the size, font codebook encapsulates richer token prior. From the results presented in \cref{codebook_size}, it can be observed that the model achieves the best performance when the codebook size is set to 1024.

\begin{table}[!ht]
\centering
%\vspace{1mm}

\begin{tabular}{ p{1.cm}<{\centering} | p{0.86cm}<{\centering}  p{0.9cm}<{\centering} p{0.9cm}<{\centering} p{0.9cm}<{\centering} p{0.9cm}<{\centering}}
\toprule
  &  L1~$\downarrow$  & RMSE~$\downarrow$ & PSNR~$\uparrow$ & SSIM~$\uparrow$ & LPIPS~$\downarrow$ \\
\midrule
128 &   0.0638&  0.214&  20.010&  0.807&0.108\\
512 &  0.0631&  0.212&  20.074&  0.809&0.107 \\
1024  &\textbf{0.0621} &\textbf{0.210}  &\textbf{20.249}   &\textbf{0.812} 
&\textbf{0.103}  \\

\bottomrule
\end{tabular}
%\caption{\textbf{Quantitative analysis of the effectiveness of the font codebook and SSEM.} C is the abbreviation for font codebook, and S for SSEM.}
\caption{\textbf{Effect of different codebook sizes when pre-training VQGAN.} 
}
\label{codebook_size}
\end{table}

\begin{figure*}[t]
\centering
\includegraphics[width=0.95\textwidth]{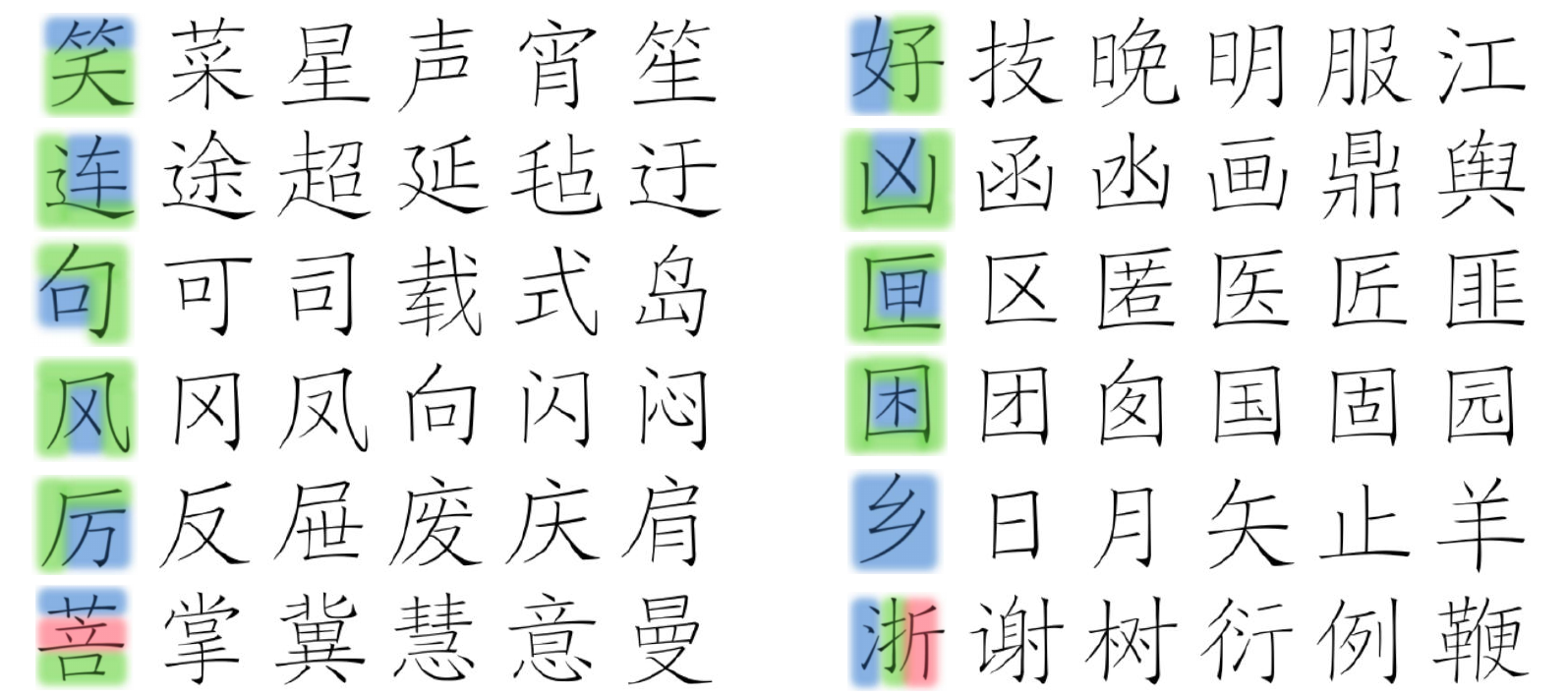} % Reduce the figure size so that it is slightly narrower than the column.
\caption{\textbf{Examples of Chinese characters with different structures.}}
\label{structure_division}
\end{figure*}

\begin{figure}[t]
\centering
\includegraphics[width=1\columnwidth]{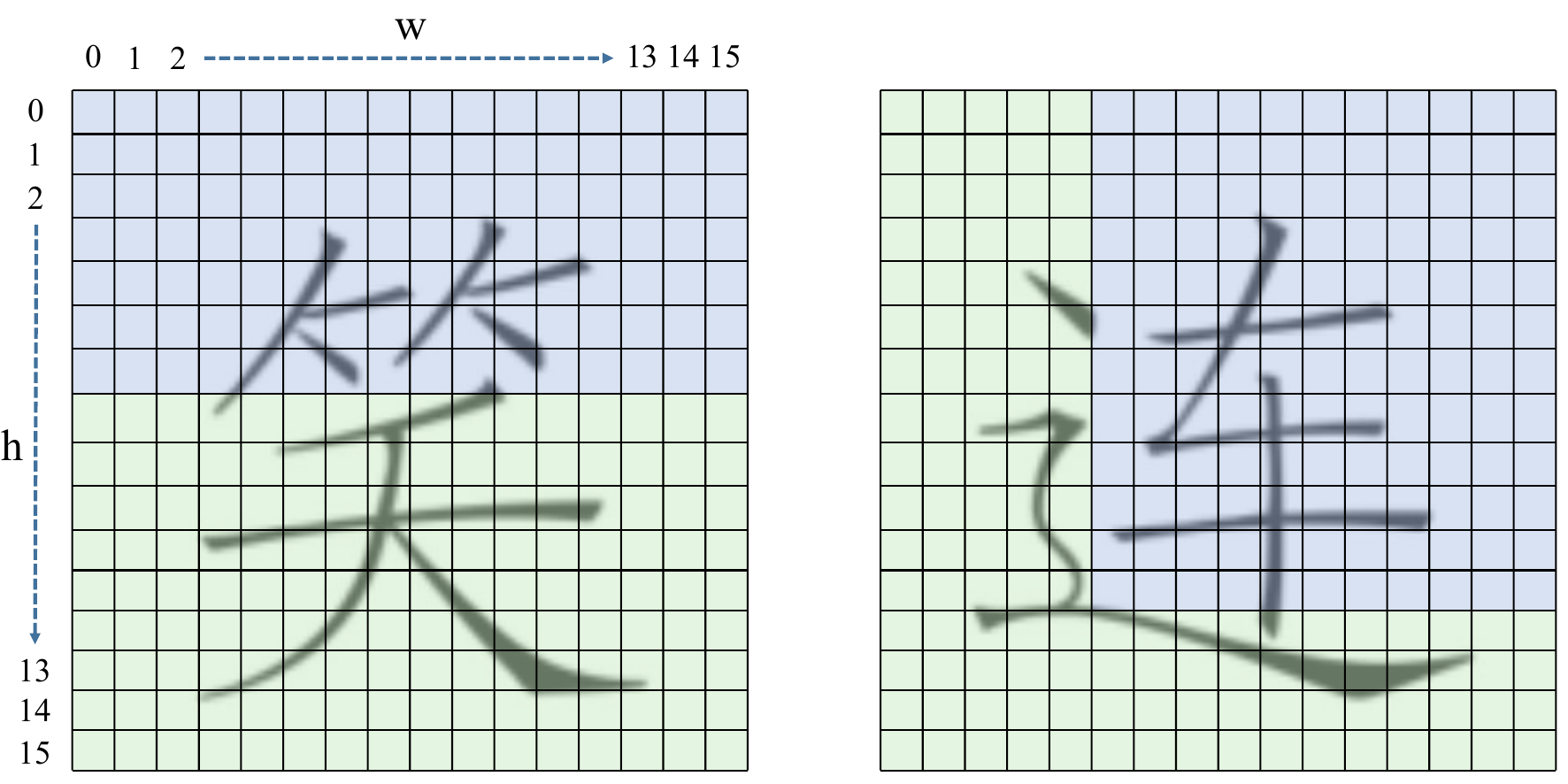} % Reduce the figure size so that it is slightly narrower than the column. Don't use precise values for figure width.This setup will avoid overfull boxes.
\caption{\textbf{Different colors represent the position sets of different structure components in the feature map.} Each small square represents a patch of the feature map.}
\label{xiao_lian}
\end{figure}

\subsection{Chinese Character Partition in SSEM}
%As illustrated in~\cref{ssrm}, we locate the positions in the content dimension $(hw)_{ci}$ and style dimension $(khw)_{si}$ of $\mA_{patch}$ that correspond to the different structure components of the content and reference glyphs, respectively. 
In the structure-level style enhancement module~(SSEM), we explicitly divide the Chinese characters with various structures to obtain the structural-level styles. Next, we will briefly describe our approach.\par
In Chinese characters, each character belongs to a specific structure. Shown in~\cref{structure_division}, Chinese characters can be divided into 12 different structure categories.
Within each structure, the structure components are arranged according to the same paradigm. This uniformity in structure allows us to explicitly decompose Chinese characters into structure components. In our experiment, we first add structure labels to each character utilized. Then, we apply the same method of division to Chinese characters that belong to the same structure.\par
Taking the left-right arrangement and bottom-left encompassed structures as examples, as shown in \cref{xiao_lian},  we partition the 16×16 feature maps into structure components based on the structure information of the characters. The remaining 10 structures are partitioned in a similar manner. Thus, we can obtain the position sets of different structure component patches. Consequently, utilizing the obtained position sets, we can locate the structure components of the content glyphs and style glyphs separately in $\mA_{\textit{patch}}\in \mathbb{R}^{hw \times khw}$, where $hw$ and $khw$ represent the resolution of the content feature maps and the style feature maps respectively. By averaging the weights in the corresponding position sets of the content and style feature maps, we can obtain structure-level attention weight $\mA_{\textit{stru}}$. 

\begin{figure*}[t]
\centering
\includegraphics[width=1\textwidth]{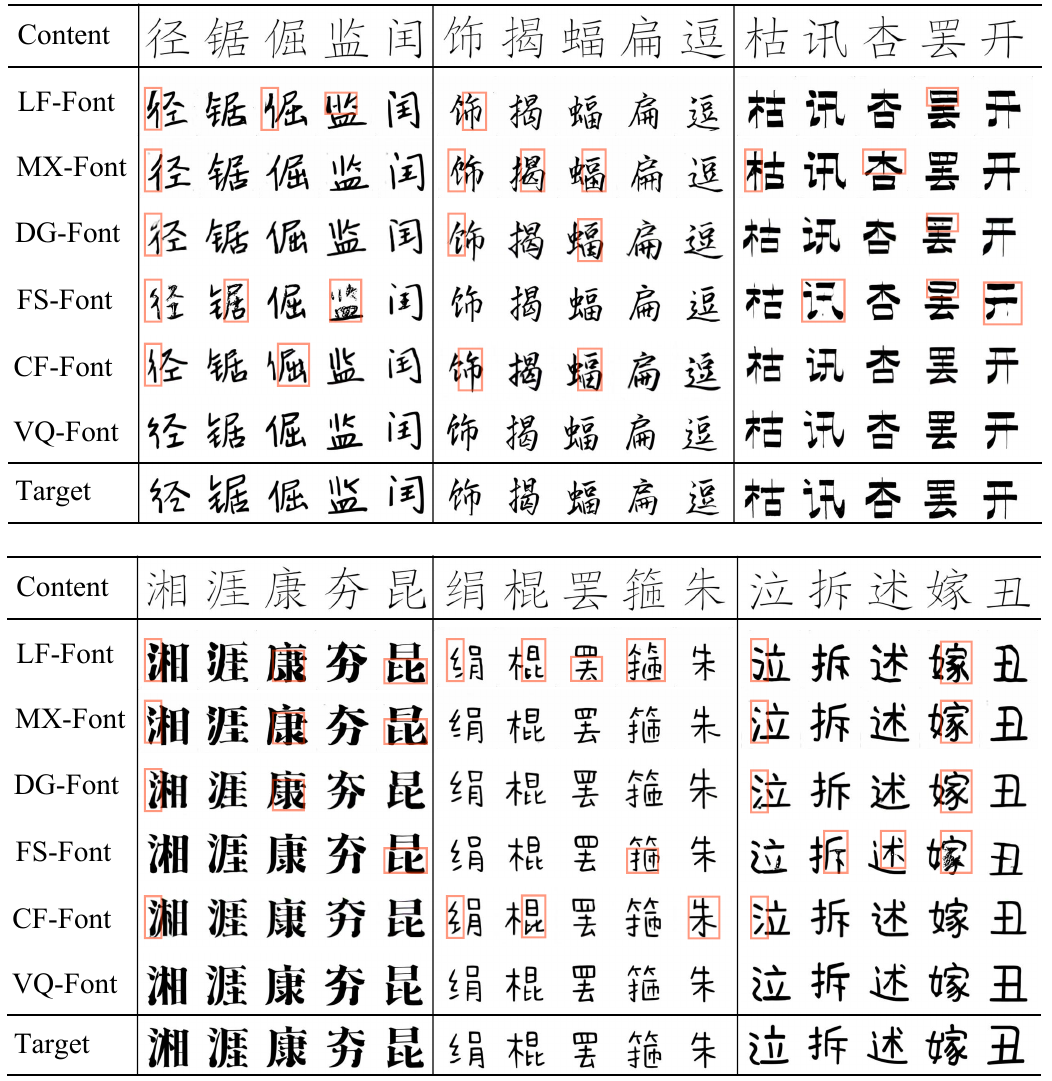} % Reduce the figure size so that it is slightly narrower than the column.
\caption{\textbf{Qualitative comparison with competing methods on SFUC dataset. 
%Results best seen at 500\% zoom.
Best view it by zooming in on the screen.}}
\label{compare_result}
\end{figure*}

\subsection{More Visual Results}
As shown in~\cref{compare_result}, we show additional comparison results with existing state-of-the-art methods~(LF-Font~\cite{park2021few}, MX-Font~\cite{park2021multiple}, DG-Font~\cite{xie2021dg}, FS-Font~\cite{tang2022few}, and CF-Font~\cite{wang2023cf}.) on SFUC dataset. From the results, it is evident that the glyphs generated by our VQ-FONT effectively address the issue of distorted strokes.\par
\begin{figure*}[!h]
\centering
\includegraphics[width=0.95\textwidth]{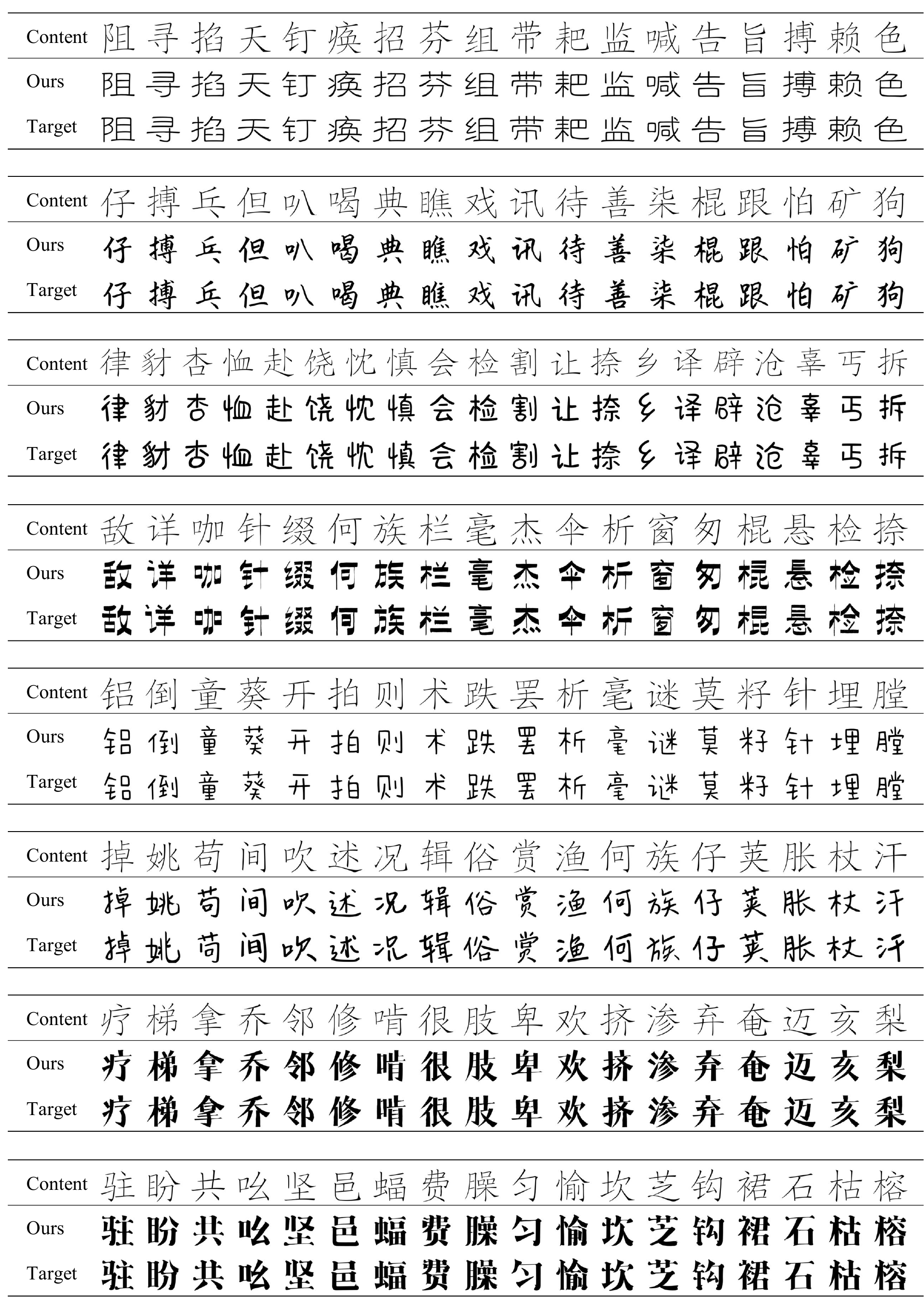} % Reduce the figure size so that it is slightly narrower than the column.
\caption{\textbf{The visual results of various fonts on SFUC dataset.}}
\label{sfuc_result}
\end{figure*}
\begin{figure*}[!h]
\centering
\includegraphics[width=0.95\textwidth]{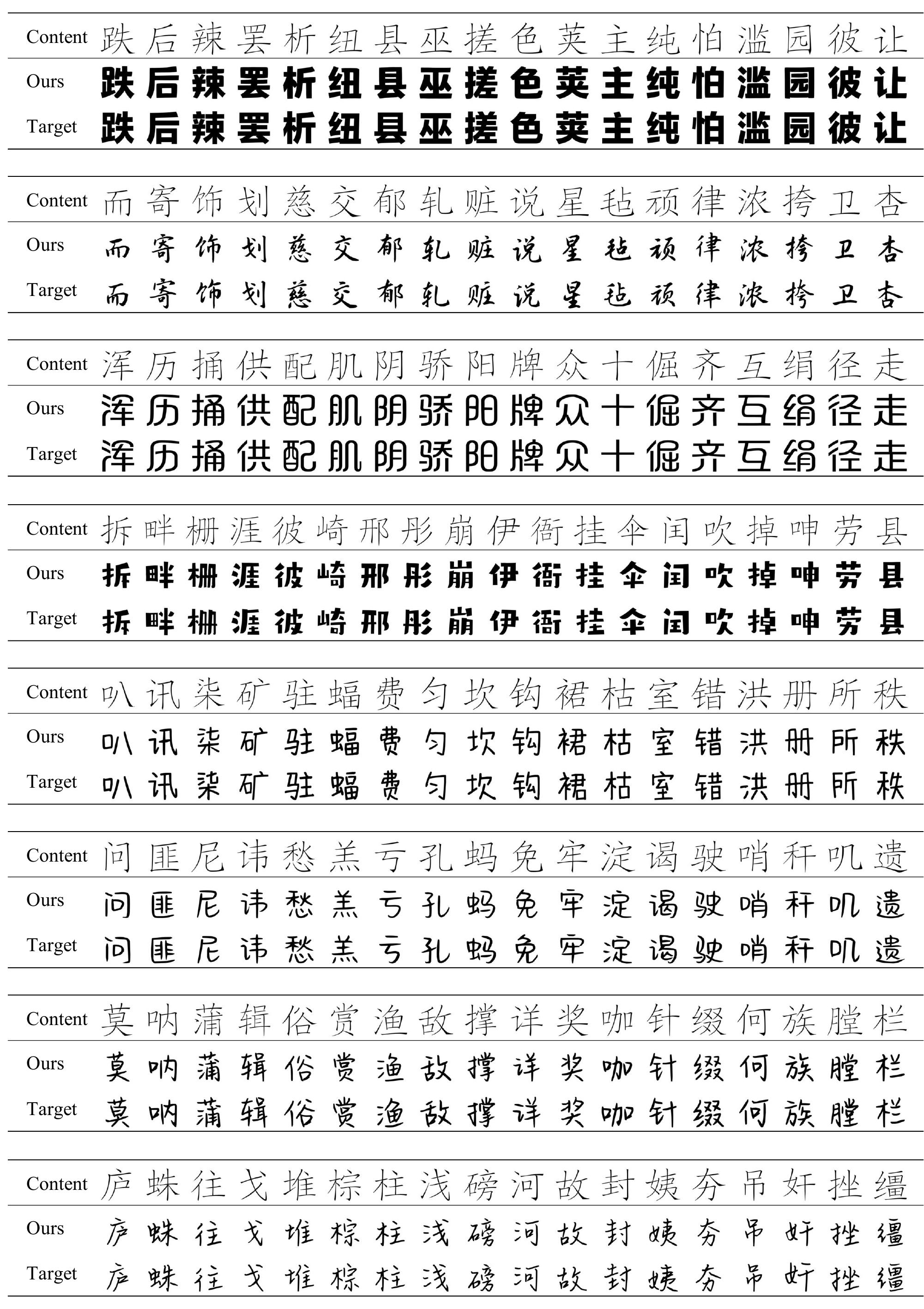} % Reduce the figure size so that it is slightly narrower than the column.
\caption{\textbf{The visual results of various fonts on UFUC dataset.}}
\label{ufuc_result}
\end{figure*}

As depicted in~\cref{sfuc_result,ufuc_result}, we present additional experimental results on the test dataset including some challenging fonts. Our model exhibits good generalization performance, allowing it to generate glyphs across a wide range of styles with impressive results.

\subsection{Examples of Different Structures}
Chinese characters can be categorized into the following structures: left-right arrangement, left-center-right arrangement, top-bottom arrangement, top-center-bottom arrangement, fully encompassed, top-three encompassed, left-three encompassed, bottom-three encompassed, top-left encompassed, top-right encompassed, bottom-left encompassed, and independent structure. As shown in~\cref{structure_division}, we provide several Chinese characters for each structure.

\end{document}